\pdfoutput=1
\documentclass[11pt]{article}

\usepackage[final]{acl}

\usepackage{times}
\usepackage{latexsym}

\usepackage[T1]{fontenc}
\usepackage[utf8]{inputenc}

\usepackage{microtype}

\usepackage{inconsolata}

\usepackage{graphicx}

\usepackage{float}
\usepackage{amsmath}
\usepackage{multirow}
\usepackage{graphicx}
\usepackage{booktabs}
\usepackage{listings}

\usepackage[utf8]{inputenc} % allow utf-8 input
\usepackage[T1]{fontenc}    % use 8-bit T1 fonts
\usepackage{hyperref}       % hyperlinks
\usepackage{url}            % simple URL typesetting
\usepackage{booktabs}       % professional-quality tables
\usepackage{amsfonts}       % blackboard math symbols
\usepackage{nicefrac}       % compact symbols for 1/2, etc.
\usepackage{microtype}      % microtypography
\usepackage{xcolor}         % colors

\title{Towards Hierarchical Multi-Step Reward Models \\for Enhanced Reasoning in Large Language Models}

\author{
  \textbf{Teng Wang\textsuperscript{1}},
  \textbf{Zhangyi Jiang\textsuperscript{1}},
  \textbf{Zhenqi He\textsuperscript{1}}, 
    \textbf{Hailei Gong\textsuperscript{2}\thanks{Corresponding author}},\\
   \textbf{Shenyang Tong\textsuperscript{1}}, 
   \textbf{Wenhan Yang\textsuperscript{1}},
     \textbf{Zeyu Li\textsuperscript{3}},
  \textbf{Yanan Zheng\textsuperscript{4}},\\
  \textbf{Zifan He\textsuperscript{1}}, 
  \textbf{Zewen Ye\textsuperscript{5}},
  \textbf{Shengjie Ma\textsuperscript{6*}}
  \textbf{Jianping Zhang\textsuperscript{7*}}
  \\
 \textsuperscript{1} the University of Hong Kong,
  \textsuperscript{2} Tsinghua University,\\
   \textsuperscript{3} Georgia Institute of Technology,
 \textsuperscript{4} National University of Singapore,\\
 \textsuperscript{5} Zhejiang University,
 \textsuperscript{6} Renmin University of China,\\
 \textsuperscript{7} the Chinese University of Hong Kong\\
  \texttt{wt0318@connect.hku.hk} 
  \\
}

\begin{document}
\maketitle

\begin{abstract}
Large Language Models (LLMs) have demonstrated strong mathematical reasoning abilities through supervised fine-tuning and reinforcement learning. 
However, existing Process Reward Models (PRMs) are vulnerable to reward hacking and require expensive, large-scale annotation of reasoning steps, limiting their reliability and scalability.
To address the first problem, we propose a novel reward model approach, \textbf{H}ierarchical \textbf{R}eward \textbf{M}odel \textbf{(HRM)}, which evaluates both individual and consecutive reasoning steps from fine-grained and coarse-grained level. HRM excels at assessing multi-step mathematical reasoning coherence, particularly in cases where a flawed step is later corrected through self-reflection.
Furthermore, to address the inefficiency of autonomously annotating PRM training data via Monte Carlo Tree Search (MCTS), we propose a lightweight data augmentation strategy, \textbf{H}ierarchical \textbf{N}ode \textbf{C}ompression (\textbf{HNC}), which merges consecutive reasoning steps within the tree structure. Applying HNC to MCTS-generated reasoning trajectories increases the diversity and robustness of HRM training data, while introducing controlled noise with minimal computational overhead.
Empirical results on the PRM800K dataset demonstrate that HRM, in conjunction with HNC, achieves superior stability and reliability in evaluation compared to PRM. Furthermore, cross-domain evaluations on MATH500 and GSM8K dataset confirm HRM’s superior generalization and robustness across diverse mathematical reasoning tasks.
The core code is publicly available \footnote{\url{https://github.com/tengwang0318/hierarchial_reward_model}}.
\end{abstract}

\section{Introduction}

\begin{table*}[]
\footnotesize
\centering
\renewcommand{\arraystretch}{1.2}

\begin{tabular}{p{4cm}|p{3cm}|p{3cm}|p{3cm}}

    \hline
    \textbf{Feature} & \textbf{ORM} & \textbf{PRM} & \textbf{HRM} \\
    \hline
    \textbf{Scoring Method} & Rule-Based or RM & RM Only & RM Only \\
    \hline
    \textbf{Granularity (Training Data)} & Whole Process & Single Step & Few Consecutive Steps \\
    \hline
    \textbf{Step-wise Feedback} & No & Yes & Yes \\
    \hline
    \textbf{Error Correction} & Yes & No & Yes \\
    \hline
\end{tabular}
\caption{Comparison of scoring methods, granularity, and feedback mechanisms across ORM, PRM, and HRM.}
    \label{tab:rm_comparison}
\end{table*}

As the scale of parameters in LLMs continues to grow~\cite{palm,gpt4,llama3,qwen25}, their general capabilities have significantly improved, surpassing human performance in various generative tasks such as text comprehension and data generation~\cite{mlprompt}. 
However, the upper bound and inherent limitations of LLMs in reasoning-intensive tasks—such as mathematical reasoning—remain an open question~\cite{gsm8k,openai_prm,google_prm,math_shepherd,google_mcts_auto_annotate,google-another-mcts,mip,mipLLM}. 
Recent approaches, such as Chain-of-Thought (CoT)~\cite{cot} and Tree-of-Thought (ToT)~\cite{tot}, have significantly enhanced reasoning performance. Despite these advancements, 
% most CoT models lack a mechanism to halt reasoning and self-reflection when an intermediate step is incorrect, leading to error propagation.
%
most CoT models lack mechanisms to detect and correct intermediate reasoning errors, resulting in continued propagation of mistakes throughout the reasoning process.
Meanwhile, ToT methods do not inherently verify every intermediate step or guarantee retrieval of the optimal reasoning trajectory, which can limit its reliability in complex problem-solving scenarios.

To mitigate these limitations, recent efforts have focused on reward mechanisms that guide LLMs effectively. 
There are two primary approaches to enhance the reasoning capabilities of LLMs from the perspective of "how to reward LLMs": the Outcome Reward Model (ORM)~\cite{openai_prm,google_prm,deepseek-r1,deepseek-math} and the Process Reward Model (PRM)~\cite{openai_prm,google_prm}. Each comes with its own limitations.
%
% ORM suffers from delayed feedback and credit assignment issues, while PRM is prone to reward hacking and incurs high costs for reasoning process annotation.
ORM suffers from delayed feedback and credit assignment issues, making it difficult to pinpoint which reasoning steps contribute to the final answer~\cite{openai_prm,google_prm}. PRM, in contrast, provides finer-grained supervision by evaluating reasoning step by step. However, most PRM methods are model-based and are prone to reward hacking~\cite{lilianweng}, where models exploit reward signals rather than genuinely improving reasoning, undermining reliability in complex tasks. Moreover, the high annotation cost associated with PRM makes large-scale deployment challenging.

In this paper, we focus on addressing the limitations of PRM. To mitigate the impact of reward hacking in PRM, we propose the \textbf{H}ierarchical \textbf{R}eward \textbf{M}odel (\textbf{HRM}). 
The term \textbf{hierarchical} highlights that, during training, HRM incorporates a hierarchical supervision signal by evaluating reasoning processes at both fine-grained (single-step) and coarse-grained (consecutive multi-step) levels. This layered approach enables HRM to capture both local and global coherence in reasoning. However, during inference, HRM remains step-wise: it assigns rewards to each reasoning step individually, the same as PRM.
Traditional PRM penalizes an incorrect single step without considering potential corrections in subsequent reasoning. In contrast, HRM assesses reasoning coherence across multiple steps, allowing the reward model to identify and incorporate later steps that rectify earlier errors, leading to a more robust and reliable evaluation.
Table~\ref{tab:rm_comparison} compares the difference between ORM, PRM and HRM. 

While HRM can be applied to other structured reasoning domains, we focus on \textit{mathematical reasoning}, the only domain with large-scale human-annotated process-level data for supervised reward modeling.
The PRM800K~\cite{openai_prm} dataset comprises manually annotated reasoning trajectories, which serve as the foundation for training ORM, PRM, and HRM. We subsequently assess the performance of Qwen2.5-72B-Math-Instruct~\cite{qwen-math} as the policy model by employing the Best-of-N Search strategy across ORM, PRM, and HRM. 
Experimental results (as shown in Table~\ref{tab:hrm_manual_prm800k}) demonstrate that HRM yields the most stable and reliable reward evaluations. 
The policy model with HRM maintains stable performance, with accuracy stabilizing at 80\% as $N$ increases. In contrast, policy models with PRM and ORM exhibit significant performance fluctuations, with accuracy degrading as $N$ grows.

% Experimental results (as shown in Table~\ref{tab:hrm_manual_prm800k}) demonstrate that HRM is the most robust to reward hacking, with accuracy stabilizing at 80\% as N increases. In contrast, PRM and ORM exhibit significant performance fluctuations, with accuracy degrading as N grows.

To fully exploit the capabilities of Monte Carlo Tree Search (MCTS) for automatic process annotation, we introduce a data augmentation framework termed \textbf{H}ierarchical \textbf{N}ode \textbf{C}ompression (\textbf{HNC}), which consolidates two consecutive nodes from different depths into a single node. This approach effectively expands the training dataset while maintaining minimal computational overhead and enhancing label robustness through controlled noise injection.
After evaluating HNC in the auto-annotation process by MCTS on the PRM800K dataset, we find that fine-tuned HRM achieves more robust scoring within PRM800K dataset and exhibits strong generalization across other domains, including GSM8K~\cite{gsm8k} and MATH500~\cite{openai_prm} dataset, outperforming PRM in robustness and consistency.

Our main contributions are as follows:

\begin{itemize}

\item We propose the \textbf{HRM}, which leverages hierarchical supervision from training data at both single-step and multi-step levels, promoting coherence and self-correction in multi-step reasoning. We validate HRM’s robustness on the PRM800K dataset using manually annotated data.

\item We introduce \textbf{HNC}, a lightweight data augmentation approach for MCTS that substantially increases the diversity and robustness of HRM training data with minimal computational cost. 
Experiments show that HRM trained on the PRM800K dataset with auto-annotated data from HNC and MCTS demonstrates improved robustness over PRM. Furthermore, HRM exhibits superior reasoning consistency and generalization across GSM8K and MATH500, consistently outperforming PRM.

\item Additionally, we enhance the policy model through fine-tuning on high-quality reasoning trajectories filtered from MCTS, further improving its reasoning performance.

    % \item We introduce \textbf{Hierarchical Node Compression (HNC)} in MCTS for autonomous annotation, refining existing reasoning trajectories by generating additional step-wise labels with controlled noise to improve score robustness. Additionally, by filtering high-quality reasoning trajectories from MCTS, we refine the policy model through fine-tuning, further enhancing its reasoning performance.

    % \item We investigate the generalization of HRM trained on PRM800K using auto-labeled reasoning processes from MCTS and HNC. Experimental results demonstrate that HRM achieves superior reasoning consistency and generalizes effectively across GSM8K and MATH500, outperforming PRM.
\end{itemize}

\begin{figure*}[t]
\centering
\includegraphics[height=0.45\textwidth]{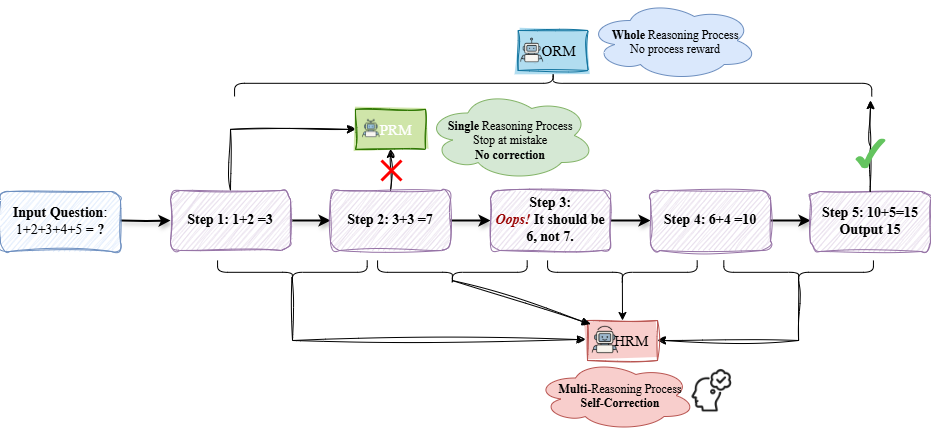}
\caption{Illustration of how ORM, PRM, and HRM handle reasoning processes. ORM evaluates the entire reasoning chain, PRM assesses individual steps but stops at errors, and HRM considers multiple consecutive steps, enabling error correction. The figure also demonstrates how HRM constructs its training dataset by merging two consecutive steps.}

\label{fig:HRM}
\end{figure*}

\section{Related Work}

\subsection{RLHF}
Reinforcement Learning with Human Feedback (RLHF)~\cite{instructGPT} is a widely used framework for optimizing LLMs by incorporating human feedback. The core idea of RLHF is to use an RM to distinguish between high-quality and low-quality responses and optimize the LLM using PPO~\cite{ppo}.

From the perspective of reward design, there are two main approaches: ORM~\cite{gsm8k,openai_prm,math_shepherd} and PRM~\cite{openai_prm,google_prm,google_mcts_auto_annotate,rest_mcts}. ORM assigns rewards based on the whole output, while PRM evaluates intermediate reasoning steps to provide more fine-grained supervision. 
% These reward mechanisms directly impact how LLMs learn to reason and optimize their outputs.

\subsection{ORM}
ORM suffers from delayed feedback and the credit assignment problem. Since rewards are only provided at the final outcome, ORM struggles to discern which intermediate steps contribute to success or failure~\cite{openai_prm}. This delayed feedback limits learning efficiency, making it harder to optimize critical decision points. Additionally, ORM is prone to spurious reasoning~\cite{gsm8k,math_shepherd}, where the model arrives at the correct answer despite flawed intermediate steps, reinforcing suboptimal reasoning patterns.
However, DeepSeek-R1\cite{deepseek-r1} integrates a rule-based ORM within GRPO\cite{deepseek-math}, demonstrating that rule-based reward, rather than score-based reward models, can effectively guide LLMs toward generating long-CoT reasoning and self-reflection, ultimately enhancing their reasoning abilities.

\begin{figure*}[t]
\centering
\includegraphics[height=0.5\textwidth]{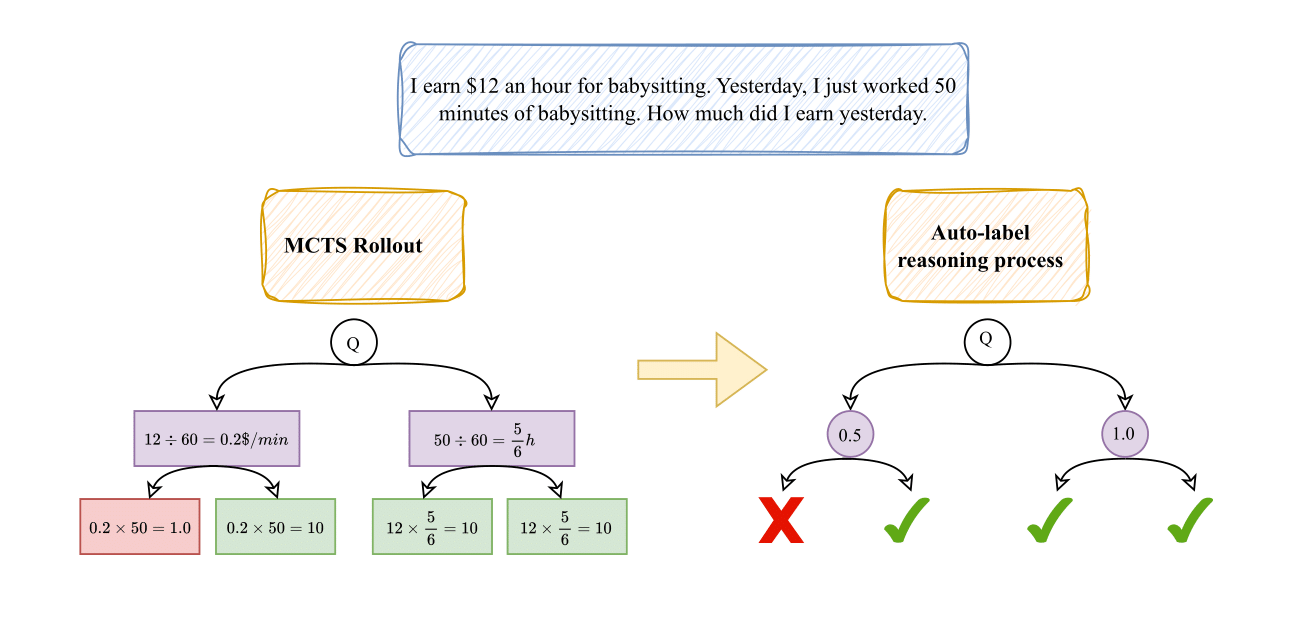}
\caption{Illustration of the MCTS-based automated reasoning annotation process. The left side depicts a tree structure where each node represents a reasoning step, simulated using the ToT approach with MCTS. The right side visualizes the assigned scores for each step in the reasoning tree.}

\label{fig:MCTS}
\end{figure*}

\subsection{PRM}
One of the most critical challenges in PRM is reward hacking, a phenomenon in which an RL agent exploits flaws or ambiguities in the reward function to achieve artificially high rewards without genuinely learning the intended task or completing it as expected ~\cite{ reward_hacking1, reward_hacking2, reward_hacking3,lilianweng}. 

Furthermore, the annotation process required for training PRM is prohibitively expensive~\cite{openai_prm,google_prm}, making large-scale implementation impractical.
To address this issue, MCTS has been proposed as an autonomous method for annotating reasoning trajectories~\cite{google_mcts_auto_annotate,rest_mcts}. 
%
% While MCTS reduces the need for human annotation, it incurs substantial computational costs due to the extensive simulations required for the MC-Score to achieve relative convergence.
%
Moreover, in MCTS, the computational cost increases significantly as the depth and breadth of the search tree expand. To mitigate this, constraints are imposed on both tree height and width, limiting the number of simulation steps and thereby reducing the diversity of generated reasoning data.

\section{Methodology}

\subsection{Hierarchical Reward Model}

PRM provides fine-grained, step-wise supervision, whereas ORM evaluates reasoning holistically. 
To combine the advantages of both, we propose the \textbf{H}ierarchical \textbf{R}eward \textbf{M}odel (\textbf{HRM}), which introduces hierarchical supervision by training on both single-step and consecutive multi-step reasoning sequences. 
HRM serves as an evaluation-oriented reward model that estimates the relative quality of reasoning trajectories, rather than directly optimizing the policy model itself. 
This hierarchical design enables HRM to capture both local accuracy and global coherence, yielding more stable and reliable reward evaluation in multi-step reasoning. 
The HRM training data extends PRM supervision by merging consecutive reasoning steps into hierarchical segments (from step 1 to $N$), as illustrated in Fig.~\ref{fig:HRM}, forming a superset of the PRM dataset.

% PRM provides fine-grained, step-wise supervision, whereas ORM evaluates reasoning holistically. To leverage the strengths of both, we propose the \textbf{H}ierarchical \textbf{R}eward \textbf{M}odel (\textbf{HRM}), which introduces hierarchical supervision by training on both single-step and consecutive multi-step reasoning sequences. This design enables HRM to capture local accuracy and global coherence, enhancing the robustness and reliability of reward evaluation in multi-step reasoning. HRM serves as an evaluation-oriented reward model that estimates the relative quality of reasoning trajectories, rather than directly optimizing the policy model itself.
% %
% The training dataset for HRM consists of consecutive reasoning sequences spanning from step 1 to $N$, as illustrated in Fig.~\ref{fig:HRM} and Section~\ref{exp:HRM}. HRM training data is a superset of PRM training data, constructed by merging consecutive reasoning steps. 

% Specifically, while PRM training data only contains individual steps with their respective reward scores, HRM training data additionally incorporates merged step pairs to enhance reasoning coherence and improve self-reflection in error correction.
%

\begin{figure*}[t]
\centering
\includegraphics[width=0.95\textwidth]{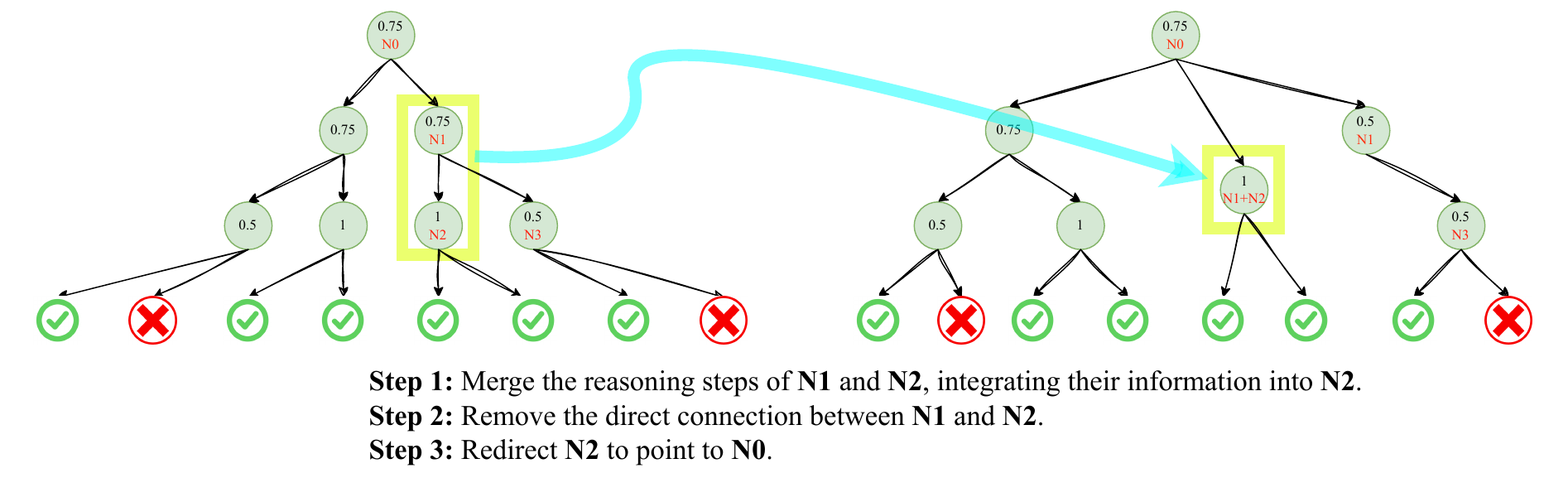}
\caption{Illustration of HNC. The left part represents the original MCTS data annotation structure, while the right part shows the transformed MCTS structure after applying HNC.}
\label{fig:HNC}
\end{figure*}

Formally, let $D$ represent the training dataset, $N$ denote the total number of reasoning steps in a sequence, $s_i$ be the $i$-th reasoning step, and $R(\cdot)$ be the reward function that assigns a score to a step. The training datasets for PRM and HRM are defined as:
% \begin{equation}
%     \begin{aligned}
%     & D_{\text{PRM}} = \{(s_i, R(s_i)) \mid 1 \leq i \leq N \}, \\
%     & D_{\text{HRM}} = D_{\text{PRM}} \cup \{(s_i + s_{i+1}, R(s_i + s_{i+1})) \mid 1 \leq i < N \}.
%     \end{aligned}
% \end{equation}

\begin{align}
&D_{\text{PRM}} = \left\{ (s_i,\, R(s_i)) \mid 1 \leq i \leq N \right\}, \\
&D_{\text{HRM}} = D_{\text{PRM}} \notag \\
&\quad \cup 
\left\{
    (s_i + s_{i+1},\, R(s_i + s_{i+1})) 
    \mid 1 \leq i < N
\right\}.
\end{align}

% Conceptually, hierarchical supervision bridges micro-level correctness (accurate individual steps) and macro-level coherence (consistent reasoning trajectories). By rewarding both local accuracy and cross-step correction, HRM encourages alignment between step-wise precision and overall reasoning integrity, analogous to compositional reasoning in cognitive frameworks.

% HRM is designed with two primary objectives: (1) capturing both fine-grained and coarse-grained reasoning consistency, and (2) enabling self-reflection and error correction. Unlike PRM, which terminates evaluation upon encountering an error, HRM assesses whether subsequent steps rectify earlier mistakes, treating them as a cohesive unit rather than isolated errors. 

Hierarchical supervision aims to connect local correctness with global coherence by jointly evaluating individual steps and their interactions. In practice, HRM achieves this through two objectives: (1) capturing both fine-grained and coarse-grained consistency, and (2) enabling self-reflection and error correction. Unlike PRM, which evaluates each step independently and penalizes early mistakes, HRM considers whether later steps can repair earlier errors, treating them as a coherent reasoning process rather than isolated faults.

% Conceptually, hierarchical supervision bridges micro-level correctness and macro-level coherence in reasoning. By rewarding both local accuracy and cross-step correction, HRM aligns step-wise precision with overall reasoning integrity, analogous to compositional reasoning in cognitive frameworks. In practice, HRM achieves this through two objectives: (1) capturing both fine-grained and coarse-grained consistency, and (2) enabling self-reflection and error correction. Unlike PRM, which evaluates each step independently and penalizes any early mistake, HRM considers whether subsequent steps repair earlier errors, treating them as a cohesive reasoning unit rather than isolated faults.

Although HRM training data incorporates merged reasoning steps, the model remains step-wise in inference, assigning a reward based solely on the current step $s_i$, similar to PRM.

This design preserves fine-grained step-level evaluation while mitigating the subjectivity of step segmentation: by merging neighboring steps during training, HRM smooths annotation noise from ambiguous boundaries and enhances robustness across reasoning granularities. Conceptually, this hierarchical supervision extends PRM in a principled way, integrating both local accuracy and global coherence. In this work, we instantiate HRM with two-step merging as the minimal yet effective form of hierarchical supervision, enabling controlled analysis while maintaining computational efficiency.

% The training dataset for HRM consists of consecutive reasoning sequences spanning from step 1 to $N$, as illustrated in Fig.~\ref{fig:HRM} and Section~\ref{exp:HRM}. HRM training data is a superset of PRM training data, constructed by merging consecutive reasoning steps. 

% This design allows HRM to learn reasoning coherence across multiple steps while still preserving the fine-grained step-wise evaluation used in PRM. By including consecutive step pairs, HRM can better capture intermediate corrections, ensuring that later steps contributing to error recovery are appropriately rewarded rather than being penalized prematurely.

\subsection{Hierarchical Node Compression in MCTS}
\label{sec:HNC}
% Although process supervision enhances the reasoning capabilities of policy models, the cost of human-annotated supervision is prohibitively high~\cite{openai_prm,google_prm}. To mitigate this, autonomous annotation methods based on MCTS have been proposed~\cite{math_shepherd, google_mcts_auto_annotate, rest_mcts}. However, these methods demand substantial computational resources, as they rely on repeatedly simulating LLM reasoning. Furthermore, ensuring the convergence of intermediate reasoning step scores requires a sufficiently deep and wide search tree; otherwise, the estimates remain biased. This exponential growth in complexity makes large-scale implementation challenging.
% Although process supervision enhances the reasoning capabilities of policy models, the cost of human-annotated supervision is prohibitively high. To address this, autonomous annotation methods based on MCTS have been proposed.  
Due to the prohibitively high cost of human-annotated supervision, autonomous annotation methods based on MCTS have been proposed.
Fig.~\ref{fig:MCTS} illustrates the process of automatic reasoning annotation using MCTS. Given a ground truth and a corresponding question, MCTS generates multiple possible reasoning paths by simulating different step-by-step solutions. Each node in the search tree represents a reasoning step, and its score is calculated based on the proportion of correct trajectories in its subtree, reflecting the likelihood that the reasoning path is valid.
However, these methods demand substantial computational resources, as achieving reliable estimates of intermediate reasoning step scores requires a sufficiently deep and wide search tree to reduce variance and mitigate bias; otherwise, the estimates may remain unreliable. This exponential growth in complexity makes large-scale implementation challenging.

To better leverage autonomous process annotation, we propose a data augmentation method called \textbf{H}ierarchical \textbf{N}ode \textbf{C}ompression (\textbf{HNC}). The key idea is to merge two consecutive nodes, each corresponding to a reasoning step, into a single node, thereby creating a new branch with minimal computational overhead. 

As shown in Fig.~\ref{fig:HNC}, HNC assumes that each node has a sufficiently large number of child nodes. By randomly  merging consecutive nodes, it introduces controlled noise, enhancing the robustness of MCTS-based scoring. 
% For instance, if each node initially has five child nodes contributing equally (20\%) to the parent node’s score, merging one child node into a new branch redistributes the weights among the remaining four, increasing their contributions to 25\%. 
Before HNC, each child node contributes \( \frac{1}{N} \) to the total score. After HNC removes a random node, the remaining child nodes redistribute their weights to \( \frac{1}{N-1} \), increasing their individual influence. Since child nodes are independent and identically distributed from the parent’s perspective, the expectation of the parent score remains unchanged. However, the variance increases from \( \frac{\sigma^2}{N} \) to \( \frac{\sigma^2}{N-1} \), introducing controlled noise that enables data augmentation at an extremely low computational cost.  
When \( N \) is sufficiently large, this variance change remains moderate while still facilitating effective data augmentation.

\subsection{Self-Training}
% After collecting reasoning data from MCTS, we filter high-quality samples using one of two approaches: the Monte Carlo Score (MC-score) or the reward model score. To mitigate reward hacking, we adapt MC-score as the primary filtering criterion. Given computational constraints, we employ supervised fine-tuning instead of reinforcement learning. The fine-tuning objective is defined as:
% \begin{equation}
%     \mathcal{L} = \mathcal{L}_{\text{LM}} + \lambda (1+\log D_{\text{KL}}(P || Q)),
% \end{equation}
% where $\mathcal{L}_{\text{LM}}$ denotes the causal language modeling loss, and $D_{\text{KL}}(P || Q)$ measures the divergence between the output distribution of the fine-tuned model and the reference model.

We filter high-quality reasoning data from MCTS using the MC-Score to ensure reliable supervision and avoid reward bias.
Due to computational constraints, we do not employ RL methods such as PPO~\cite{ppo} or GRPO~\cite{deepseek-math}. Instead, we continue using supervised fine-tuning. To preserve the general capabilities of the policy model, we incorporate causal language modeling loss combined with KL divergence regularization using a reference model. The objective function is defined as:

\begin{equation}
    \mathcal{L} = \mathcal{L}_{\text{LM}} + \lambda \log D_{\text{KL}}(P || Q),
\end{equation}

\label{method:self-train}

where $\mathcal{L}_{\text{LM}}$ represents the causal language modeling loss computed on high-quality reasoning sequences, and $D_{\text{KL}}(P || Q)$ denotes the KL divergence between the policy model's output distribution $P$ and the reference model's output distribution $Q$. The term $\lambda$ serves as a weighting factor to balance task-specific adaptation and retention of general capabilities.

\begin{table}[t]
\footnotesize
\centering
\renewcommand{\arraystretch}{1.2}

\begin{tabular}{c|c|c|c|c|c}
    \hline
    \textbf{N} & 2 & 4 & 8 & 16 & 24 \\
    \hline
    \textbf{ORM} & 0.622 & 0.677&  0.655&  0.655 & 0.633
\\
    \hline
    \textbf{PRM} & 0.700 & 0.644 & 0.611 & 0.588 & 0.577
\\
    \hline
    \textbf{HRM} & \textbf{0.722} &   \textbf{0.711} &  \textbf{0.744} &  \textbf{0.800} &  \textbf{0.800}\\
    \hline

\end{tabular}
 % \caption{Accuracy of Qwen2.5-72B-Math-Instruct evaluated under the best-of-$N$ strategy using ORM, PRM, and HRM on the PRM800K test set. ORM, PRM, and HRM are fine-tuned on the manually labeled PRM800K dataset.}
 \caption{Accuracy of Qwen2.5-72B-Math-Instruct on the PRM800K test set under the best-of-$N$ strategy, comparing ORM, PRM, and HRM. All models are fine-tuned on the manually labeled PRM800K dataset.}

    \label{tab:hrm_manual_prm800k}
\end{table}

\setlength{\tabcolsep}{3pt}
\begin{table*}[t]
\centering
\footnotesize
\renewcommand{\arraystretch}{1.2}

\begin{tabular}{c|c|cccccccccc}
    \toprule
    \multirow{2}{*}{\textbf{Policy Model}} & \multirow{2}{*}{\textbf{Method}} & \multicolumn{10}{c}{\textbf{N}} \\
    \cmidrule(lr){3-12}
    &  & 2 & 4 & 8 & 16 & 24 & 32 & 64 & 128 & 256 & 512 \\
    \midrule

    \multirow{2}{*}{DeepSeek-Math-7B} 
    & PRM  & 0.311 & 0.433 & 0.377 & 0.455 & 0.411 & 0.455 & 0.466 & 0.444 & 0.377 & 0.377 \\
    & HRM  & 0.311 & 0.388 & \textbf{0.444} & 0.455 & \textbf{0.455} & 0.422 & \textbf{0.533} & \textbf{0.522} & \textbf{0.455} & \textbf{0.500} \\
    \midrule

    \multirow{2}{*}{Qwen2.5-72B-Math} 
    & PRM  & 0.233 & 0.344 & 0.411 & 0.422 & 0.488 & 0.522 & 0.600 & 0.566 & 0.666 & 0.700 \\
    & HRM  & \textbf{0.288} & \textbf{0.366} & 0.366 & \textbf{0.488} & \textbf{0.511} & \textbf{0.611} & \textbf{0.622} & \textbf{0.611} & \textbf{0.711} & \textbf{0.722} \\
    \midrule

    \multirow{2}{*}{Qwen2.5-7B-Math} 
    & PRM  & 0.477 & 0.466 & 0.600 & 0.544 & 0.633 & 0.677 & 0.733 & 0.677 & 0.700 & 0.722 \\
    & HRM  & \textbf{0.500} &\textbf{ 0.566} &\textbf{ 0.655} & \textbf{0.600} & \textbf{0.666} & \textbf{0.711}& 0.711 & \textbf{0.766} & \textbf{0.777} & \textbf{0.766} \\
    \bottomrule
\end{tabular}

\caption{Accuracy of different policy models under PRM and HRM using the best-of-$N$ strategy on the PRM800K test set. The training data for both PRM and HRM are derived from MCTS with Qwen2.5-7B-Math-Instruct.}
\label{tab:hrm_mcts_prm800k}
\end{table*}

Without proper KL regularization or with an insufficiently weighted KL loss (i.e., a very small $\lambda$), the KL divergence grows unbounded during training. Specifically, KL loss typically ranges from $0$ to $20000$, whereas the causal LM loss remains within $0$ to $12$, leading to a severe loss imbalance. This causes the optimization process to excessively minimize KL divergence at the expense of task-specific reasoning performance.
To address this, we apply a logarithmic scaling to $D_{\text{KL}}(P || Q)$, stabilizing the loss landscape and ensuring a balanced trade-off between preserving general language capabilities and enhancing reasoning ability. Further details are provided in Section~\ref{exp:self-training}.

\section{Experiment}
\subsection{HRM}
\label{exp:HRM}

Given that the PRM800K dataset consists of Phase1 and Phase2, where Phase1 includes manually annotated reasoning processes, we utilize these manual annotations to construct the training datasets for ORM, PRM, and HRM.  
ORM training data comprises complete reasoning trajectories, while PRM training data consists of individual reasoning steps conditioned on preceding context. HRM training data extends PRM by incorporating multiple consecutive reasoning steps, allowing HRM to capture self-reflection and ensure reasoning coherence across sequential steps.
Table~\ref{tab:manual_label} summarizes the labeling rules for merged reasoning steps in HRM.

We fine-tune Qwen2.5-1.5B-Math~\cite{qwen-math} as the RM for classifying given reasoning step as correct or incorrect. Given an input, RM predicts logits for the \textit{positive} and \textit{negative} classes, denoted as $l_{\text{pos}}$ and $l_{\text{neg}}$, respectively. The score is obtained by applying the softmax function:

\begin{equation}
    P(y = \text{pos} \mid x) = \frac{\exp(l_{\text{pos}})}{\exp(l_{\text{pos}}) + \exp(l_{\text{neg}})},
\end{equation}

where $P(y = \text{pos} \mid x)$ denotes the probability of given reasoning step being correct. This probability serves as the reward assigned by RM.  
Detailed information is provided in Appendix~\ref{appendix:HRM_train}.

To evaluate the performance of ORM, PRM, and HRM, we employ Qwen2.5-72B-Math-Instruct~\cite{qwen-math} as the policy model and implement the best-of-$N$ strategy. Specifically, ORM selects the best result from $N$ complete reasoning trajectories, while PRM and HRM score $N$ intermediate reasoning steps and select the most promising one at each step. 
For PRM and HRM, we consider the completion of a formula as an intermediate reasoning step, enabling a finer-grained evaluation mechanism.
%
% Table~\ref{tab:hrm_manual_prm800k} presents the results, showing that ORM and PRM exhibit significant fluctuations, with accuracy decreasing as $N$ increases. In contrast, HRM maintains stable performance, converging to an accuracy of 80\% as $N$ grows. 
Table~\ref{tab:hrm_manual_prm800k} presents the results, showing that the accuracy of the policy model with ORM and PRM exhibits significant fluctuations, decreasing as $N$ increases. In contrast, the policy model with HRM maintains stable performance, converging to an accuracy of 80\% as $N$ grows.

% \paragraph{Statistical robustness.}
% The evaluation itself already provides extensive statistical averaging.
% Each dataset contains thousands of test problems (1k in GSM8K, 500 in MATH500, and over 800k annotated reasoning steps in PRM800K), which smooths random variation across samples.
% Moreover, in the Best-of-$N$ evaluation (up to $N=512$), each reasoning step expands into $N$ candidate paths, forming an exponentially large search tree.
% This hierarchical exploration implicitly averages over a vast number of trajectories, effectively functioning as multi-seed sampling and significantly reducing evaluation variance.
\paragraph{Statistical robustness.}
The evaluation itself already provides extensive statistical averaging.
Each dataset contains thousands of test problems (1k in GSM8K, 500 in MATH500, and over 800k annotated reasoning steps in PRM800K), which smooths random variation across samples.
Notably, MATH500 and PRM800K consist of challenging high-school and university-level problems (e.g., calculus, linear algebra), whose solutions require multi-step reasoning rather than surface-level pattern matching or guessing.
Moreover, in the Best-of-$N$ evaluation (up to $N=512$), each reasoning step expands into $N$ candidate paths, forming an exponentially large search tree.
This hierarchical exploration implicitly averages over a vast number of trajectories, effectively functioning as multi-seed sampling and significantly reducing evaluation variance.

% \subsection{HNC}
% In this section, we utilize only the questions and ground truth from the PRM800K dataset~\cite{openai_prm}, without relying on manually annotated data. We employ MCTS with Qwen2.5-7B-Math-Instruct~\cite{qwen25} to simulate and generate multiple reasoning trajectories within a tree structure.  
% %
% Each node in the tree is scored based on the proportion of correct reasoning steps among all steps in its subtree, where the node serves as the root.
% %
% The resulting reasoning trajectories are used to train PRM, after which we apply the HNC data augmentation method (Section~\ref{sec:HNC}) to generate additional training data for HRM.  

\begin{figure*}[t]
\centering
\includegraphics[width=0.8\textwidth]{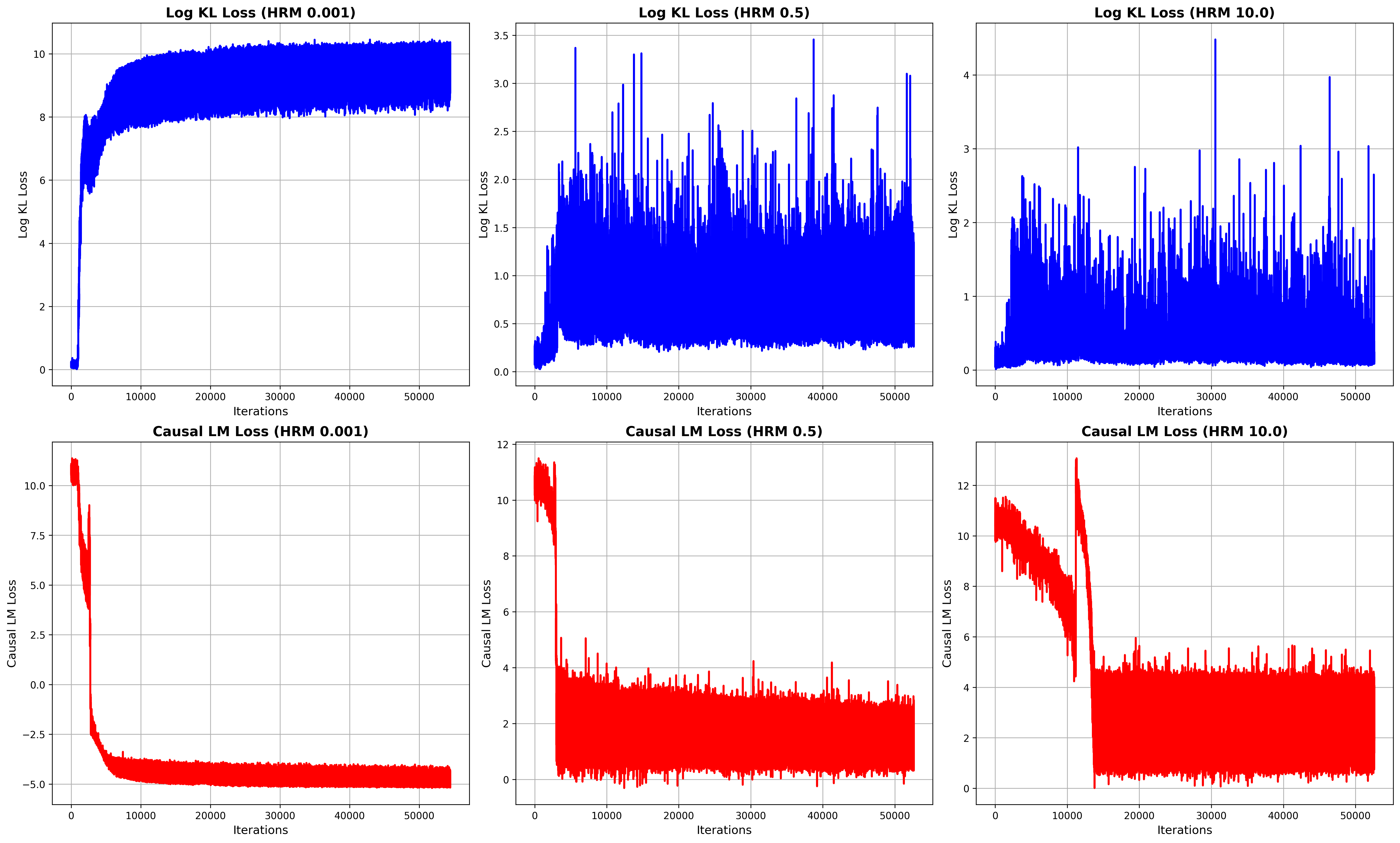}
\caption{Loss dynamics during training across different KL loss weightings. Each column corresponds to a different $\lambda$ value: 0.001 (left), 0.5 (middle), and 10.0 (right). The top row shows the log KL loss, while the bottom row depicts the causal language modeling loss. 
% A small $\lambda$ (0.001) results in an excessively large KL loss, preserving general ability but potentially compromising task-specific performance. Conversely, a large $\lambda$ (10.0) limits improvements in reasoning ability and slows convergence.
}

\label{fig:KL}
\end{figure*}

\subsection{HNC}
\label{HNC}
In this section, we utilize only the questions and ground truth from the PRM800K dataset, without relying on manually annotated data. 
%We employ MCTS with Qwen2.5-7B-Math-Instruct~\cite{qwen-math} as an automatic annotation method to generate reasoning trajectories.  
We adopt MCTS as the automatic annotation method, using Qwen2.5-7B-Math-Instruct as the reasoning engine to generate trajectories.
As mentioned in Section~\ref{sec:HNC}, these auto-annotated reasoning trajectories from MCTS are used to train PRM, after which we apply the HNC data augmentation method to generate additional training data for HRM.

To balance computational efficiency and robustness, we configure MCTS with 5–6 child nodes per parent and a maximum tree depth of 7, ensuring reasoning completion within 7 steps. Since the computational cost of MCTS rollouts grows exponentially with tree depth and branching factor, we limit these parameters to maintain feasibility. The full MCTS simulation requires approximately 2,457 A100 GPU-hours, while the HNC augmentation process takes around 30 minutes with CPU.
% To balance computational efficiency and robustness, we configure MCTS with 5–6 child nodes per parent and a maximum tree depth of 7, ensuring reasoning completion within 7 steps. The full MCTS simulation requires approximately 2,457 A100 GPU-hours, while the HNC augmentation process takes around 30 minutes. 
%

We perform supervised fine-tuning of Qwen2.5-1.5B-Math for both PRM and HRM. Training HRM adds roughly 30\% more samples compared to PRM, increasing total training time from about 84 GPU hours to 120 GPU hours on A100 GPUs. This additional cost is negligible relative to the over 2400 GPU hours required for generating the reasoning trajectories via MCTS, which constitute the true computational bottleneck.

 To evaluate performance, we employ different policy models, including Qwen2.5-7B-Math-Instruct~\cite{qwen-math}, DeepSeek-Math-7B~\cite{deepseek-math}, and Qwen2.5-72B-Math-Instruct~\cite{qwen-math}, applying the best-of-$N$ strategy on the PRM800K dataset.  
Detailed training information is provided in Appendix~\ref{appendix:HNC}.

\setlength{\tabcolsep}{3pt}
\begin{table*}[t]
\centering
\footnotesize
\renewcommand{\arraystretch}{1.2}

\begin{tabular}{c|c|cccccccccc}
    \toprule
    \multirow{2}{*}{\textbf{Policy Model}} & \multirow{2}{*}{\textbf{Method}} & \multicolumn{10}{c}{\textbf{N}} \\
    \cmidrule(lr){3-12}
    &  & 2 & 4 & 8 & 16 & 24 & 32 & 64 & 128 & 256 & 512 \\
    \midrule

    \multirow{2}{*}{Qwen2.5-7B-Math} 
    & PRM  & 0.477 & 0.466 & 0.600 & 0.544 & 0.633 & 0.677 & 0.733 & 0.677 & 0.700 & 0.722 \\
    & HRM  & \textbf{0.500} &\textbf{ 0.566} &\textbf{ 0.655} & \textbf{0.600} & \textbf{0.666} & \textbf{0.711}& 0.711 & \textbf{0.766} & \textbf{0.777} & \textbf{0.766} \\
    \midrule

    \multirow{2}{*}{Qwen-7B-PRM} 
    & PRM  & 0.477 & 0.555 & 0.533& 0.655& 0.655& 0.677& 0.711& 0.700& 0.744& 0.744\\
    & HRM   & 0.477 & 0.544 & \textbf{0.644} & \textbf{0.722} & \textbf{0.700} & \textbf{0.711} & \textbf{0.722} & \textbf{0.755} & \textbf{0.800} & \textbf{0.800}  \\
    \midrule

    \multirow{2}{*}{Qwen-7B-HRM} 
    & PRM  & 0.511 & 0.533 & 0.644 & 0.667 & 0.667& 0.689 & 0.733 & 0.722 & 0.744 & 0.733
 \\
    & HRM  & 0.489 & \textbf{0.589} & \textbf{0.722} &  \textbf{0.722} &  \textbf{0.722}&  \textbf{0.744}&  \textbf{0.744} & \textbf{0.789} & \textbf{0.789} & \textbf{0.789}
    \\
    \bottomrule
\end{tabular}

% \caption{
% Comparison of fine-tuned policy model reasoning performance using high-quality reasoning data derived from MCTS. 
% % We perform SFT using high-MC-score reasoning data from PRM/HRM training datasets. 
% Qwen-7B-HRM refers to the policy model obtained by fine-tuning Qwen2.5-7B-Math on high-MC-score reasoning data from HRM's training set, while Qwen-7B-PRM follows the same procedure with PRM's training set. 
% % The results demonstrate that SFT enhances the reasoning capabilities of the policy model, with HRM exhibiting greater robustness compared to PRM.
% }

\caption{
Comparison of fine-tuned policy model reasoning performance on the PRM800K dataset using the best-of-$N$ strategy.  
Qwen-7B-HRM denotes the policy model fine-tuned on high-MC-score reasoning data from HRM's training set, while Qwen-7B-PRM follows the same procedure for PRM's training set.  
}

\label{tab:hrm_self_training}
\end{table*}

\setlength{\tabcolsep}{3pt}
\begin{table*}[t]
\centering
\footnotesize
\renewcommand{\arraystretch}{1.2}

\begin{tabular}{c|c|cccccccccc}
    \toprule
    \multirow{2}{*}{\textbf{Dataset}} & \multirow{2}{*}{\textbf{Method}} & \multicolumn{10}{c}{\textbf{N}} \\
    \cmidrule(lr){3-12}
    &  & 2 & 4 & 8 & 16 & 24 & 32 & 64 & 128 & 256 & 512 \\
    \midrule

    \multirow{2}{*}{GSM8K} 
    & PRM & 0.784 & 0.828 & 0.858 & 0.882 & 0.884 & 0.893 &  0.905 & 0.917 & 0.927 & 0.918 \\
    & HRM &  0.784 &  \textbf{0.833} &  0.846& \textbf{0.886} &  \textbf{0.893}& \textbf{0.902}& \textbf{0.907}&  0.914& \textbf{0.930} & \textbf{0.926}
 \\
    \midrule

    \multirow{2}{*}{MATH500} 
    & PRM  & 0.468 & 0.572 & 0.598 &  0.624 & 0.658 &  0.658& 0.656& 0.662& 0.686& 0.688
 \\
    & HRM  & \textbf{0.490} & \textbf{0.576} &  \textbf{0.612} &  \textbf{0.660} &  \textbf{0.688} & \textbf{0.692} &  \textbf{0.742} &  \textbf{0.740}& \textbf{0.740} & \textbf{0.736}
 \\
   
    \bottomrule
\end{tabular}
\caption{Generalization performance of PRM and HRM trained on PRM800K and evaluated on GSM8K and MATH500 under the Best-of-$N$ setting using Qwen2.5-7B-Math-Instruct.}

\label{tab:hrm_different_domain}
\end{table*}

Table~\ref{tab:hrm_mcts_prm800k} presents the accuracy results of various policy models under PRM and HRM settings on the PRM800K dataset. Although both PRM and HRM training data are derived from MCTS with Qwen2.5-7B-Math-Instruct, we evaluate HRM and PRM using different policy models, where HRM consistently demonstrates greater stability and robustness compared to PRM.

% The relatively lower performance of Qwen2.5-72B-Math-Instruct can be attributed to the tree height constraints imposed by MCTS, which necessitate answer generation within a predefined template and a fixed number of steps. Although Qwen2.5-72B-Math-Instruct demonstrates strong reasoning capabilities, its highly structured training process renders it susceptible to performance degradation when deviating from its learned format. To ensure computational feasibility, our MCTS framework enforces limitations on tree depth, which further amplifies this effect.
% The relatively lower performance of Qwen2.5-72B-Math-Instruct (as shown in Table~\ref{tab:hrm_mcts_prm800k}) can be attributed to the tree height constraints imposed by MCTS, which limit answer generation to a predefined template of at most 6 reasoning steps and require explicit output of "\# Step X".
% While Qwen2.5-72B-Math-Instruct exhibits strong reasoning capabilities, its highly structured training process makes it more vulnerable to performance degradation when deviating from its learned format.
The relatively lower performance of Qwen2.5-72B-Math-Instruct (as shown in Table~\ref{tab:hrm_mcts_prm800k}) can be attributed to the tree height constraints imposed by MCTS, which limit answer generation to a predefined template of at most 6 reasoning steps and require explicit output of "\# Step X". While Qwen2.5-72B-Math-Instruct exhibits strong reasoning capabilities, its highly structured training process makes it more likely to generate outputs that deviate from the required format. As a result, some outputs cannot be retrieved using regex-based post-processing, thereby affecting the overall measured performance.

\subsection{Self-Training}
\label{exp:self-training}

We adapt the method described in Section~\ref{method:self-train} to filter high-quality reasoning data and train the policy model.
Fig.~\ref{fig:KL} illustrates that when $\lambda$ is small (e.g., 0.001), the fine-tuned policy model rapidly loses its generalization ability within just a few iterations, causing the KL loss to escalate to approximately 20,000. In contrast, the causal LM loss remains within the range of 0 to 12, leading to a significant imbalance. This discrepancy underscores the necessity of applying logarithmic scaling to the KL term in the objective function, as discussed in Section~\ref{method:self-train}.  
Conversely, when $\lambda$ is excessively large (e.g., 10.0), the model prioritizes adherence to the reference distribution, resulting in slower convergence and constrained improvements in reasoning capability.

% We fine-tune Qwen2.5-7B-Math-Instruct using high-MC-score reasoning data extracted from the PRM and HRM training datasets. Qwen-7B-HRM denotes the policy model fine-tuned on high-MC-score reasoning data from HRM's training set, while Qwen-7B-PRM follows the same procedure for PRM's training set. We set $\lambda$ to 0.5.
We fine-tune Qwen2.5-7B-Math-Instruct using reasoning data with MC-score $> 0.9$, extracted from the PRM and HRM training datasets. Qwen-7B-HRM denotes the policy model fine-tuned on such data from HRM's training set, while Qwen-7B-PRM follows the same procedure for PRM's training set. We set $\lambda$ to 0.5.
Table~\ref{tab:hrm_self_training} further validates that SFT enhances the policy model’s reasoning capability by leveraging high-quality data, with HRM demonstrating greater robustness compared to PRM.

\subsection{HRM Generalization Across Different Domains}

% To broaden the applicability of HRM and evaluate its generalization capability, we assess HRM and PRM, trained on the PRM800K dataset, on the Math500~\cite{openai_prm} and GSM8K~\cite{gsm8k} datasets. Table~\ref{tab:hrm_different_domain} shows that HRM demonstrates superior generalization in Math500 compared to PRM, indicating its ability to handle complex mathematical reasoning tasks. 
% To broaden the applicability of HRM and evaluate its generalization capability, we assess HRM and PRM, trained on the PRM800K dataset, on the Math500~\cite{openai_prm} and GSM8K~\cite{gsm8k} datasets.
Trained solely on PRM800K, HRM and PRM are further evaluated on MATH500~\cite{openai_prm} and GSM8K~\cite{gsm8k} to assess their generalization to out-of-domain reasoning tasks.
Table~\ref{tab:hrm_different_domain} shows that HRM exhibits greater robustness across different domains, demonstrating superior generalization performance, particularly in MATH500, where it effectively handles complex mathematical reasoning tasks.

However, the performance difference between HRM and PRM on the GSM8K dataset is marginal, as GSM8K primarily consists of relatively simple arithmetic problems.  
A strong policy model can typically solve these problems within three steps, reducing the impact of HRM’s key advantages, such as assessing multi-step reasoning coherence and facilitating self-reflection.  
Nevertheless, as shown in Table~\ref{tab:hrm_different_domain}, HRM still achieves a slight performance edge over PRM, even on simpler datasets like GSM8K.

% However, the performance difference between HRM and PRM in GSM8K dataset is marginal, as the GSM8K dataset primarily consists of relatively simple arithmetic problems. 
% %
% A strong policy model can typically solve these problems within three steps on the GSM8K dataset, limiting the impact of HRM’s advantages, such as its ability to assess multi-step reasoning coherence and facilitate self-reflection.
% %
% Nevertheless, as shown in Table~\ref{tab:hrm_different_domain}, HRM still slightly outperforms PRM on simpler datasets like GSM8K. 

% More importantly, in more complex datasets such as Math500, HRM demonstrates significantly greater robustness compared to PRM, highlighting its effectiveness in handling challenging reasoning tasks.

\section{Conclusion}

In this paper, we present HRM, which enhances multi-step reasoning evaluation by integrating fine-grained and coarse-grained assessments, improving reasoning coherence and self-reflection.  
We further introduce HNC, a data augmentation method that optimizes MCTS-based autonomous annotation, enhancing label diversity while expanding training data with minimal computational cost.  
Extensive experiments on PRM800K dataset demonstrate HRM's superior robustness over PRM, with strong generalization across GSM8K and MATH500 dataset. Additionally, MCTS-generated auto-labeled data enables policy model fine-tuning, further improving reasoning performance.

\section{Limitations}
\subsection{Merged Steps}

The reason why we only merge two consecutive steps at a time is to maintain the simplicity and clarity of the labeling strategy. When merging more than two steps—such as combining four reasoning steps—the number of possible label combinations increases rapidly (e.g., one positive, two negative, one neutral), making it difficult to define unified labeling rules and leading to potential conflicts. This stands in sharp contrast to the straightforward rules shown in Table~\ref{tab:manual_label}, where merging only two steps allows for clear and consistent label definitions.

\bibliography{custom}

\appendix
\clearpage
\section{Appendix}

\subsection{Labeling Rules for HRM Training Data}

Table~\ref{tab:manual_label} summarizes the labeling rules for merging reasoning steps in HRM, as applied to the manually labeled PRM800K dataset. 
We note that labeling consecutive neutral steps as negative is a deliberate design choice, motivated by the observation that such sequences often indicate non-progressive reasoning and inefficiency, which we aim to penalize to encourage concise and goal-directed reasoning.

\begin{table*}[]
\footnotesize
\centering
\renewcommand{\arraystretch}{1.2}

\begin{tabular}{c|c|c}
% \begin{tabular}{p{2cm}|p{2cm}|p{1.6cm}}

    \hline
    Previous Step Label & Current Step  Label & \textbf{Label for Merged Step} \\
    \hline
    Positive & Positive & \textbf{Positive}  \\
    \hline
    Positive & Neutral/Negative & \textbf{Negative}  \\
    \hline
    Neutral/Negative & Positive & \textbf{Positive}  \\
    \hline
    Neutral/Negative & Neutral/Negative & \textbf{Negative} \\
    \hline
\end{tabular}
\caption{Labeling strategy for constructing the HRM training dataset from manual annotations in PRM800K dataset. PRM800K dataset contains three label types: Positive, Negative, and Neutral. HRM extends PRM by incorporating multi-step reasoning.}
\label{tab:manual_label}
\end{table*}

\subsection{HRM Training Details}
\label{appendix:HRM_train}

To accelerate the training process of reward model, we employ FlashAttention~\cite{flashattention,flashattention2}, DeepSpeed~\cite{deepspeed1,deepspeed2}, and mixed-precision training~\cite{bf16,fp16}. However, within the PRM800K domain, we frequently encounter the issue: \textit{"Current loss scale already at minimum - cannot decrease scale anymore. Exiting run."} This indicates that the numerical precision is insufficient for stable training. To mitigate this issue and ensure reproducibility, we set \texttt{max\_grad\_norm} to 0.01, which effectively stabilizes the training process.

We define the completion of a reasoning step as the end of a formula, using stop = \verb|['\]\n\n', '\)\n\n', '# END!']| as boundary markers.

The following prompt is used in Section~\ref{exp:HRM}:
\begin{lstlisting}[basicstyle=\footnotesize\ttfamily, breaklines=true, frame=single]
"""
You are an expert of Math and need to 
solve the following question and return 
the answer.
    
Question:
{question}

Let's analyze this step by step.

After you finish thinking, you need to 
output the answer again! 

The answer should start with '# Answer', 
followed by two line breaks and the 
final response.

Just provide the answer value without 
any descriptive text at the end. 

And the answer ends with '# END!'
Below is a correct example of the 
expected output format:
-----------------
Question: 1+2+3 = ?

Firstly, solve 1 + 2 = 3,
Then, 3 + 3 = 6.

# Answer  

6
# END!
-----------------
"""
\end{lstlisting}

\subsection{HNC Setting Details}
\label{appendix:HNC}
To ensure the feasibility of autonomous annotation using MCTS, we impose constraints on both the width and height of the search tree. This limitation prevents us from treating the completion of a formula as a single reasoning step. Instead, we require the model to explicitly output \texttt{\# Step X} at each step. Consequently, the training data for the reward model is segmented using \texttt{\# Step X} as a delimiter. During inference, we also apply \texttt{\# Step X} as a separator and employ the Best-of-N strategy for selecting the optimal reasoning path.

The prompt we use is as follows(delimiter=\texttt{['\# END!', '\# Step 2', "\# Step 3", "\# Step 4", "\# Step 5"]}):
\begin{lstlisting}[basicstyle=\footnotesize\ttfamily, breaklines=true, frame=single]
"""You are an expert of Math and need to 
solve the following question and return 
the answer.

Question:
{question}


Let's analyze this step by step.

Begin each step with '# Step X' to 
clearly indicate the entire reasoning 
step.

After you finish thinking, you need to 
output the answer again! 

The answer should start with '# Answer', 
followed by two line breaks and the 
final response.

Just provide the answer value without 
any descriptive text at the end. 

And the answer ends with '# END!'

Below is a correct example of the 
expected output format:
-----------------
Question: 1+2+3 = ?

# Step 1
solve 1 + 2 = 3,

# Step 2
Then, 3 + 3 = 6.

# Answer  

6
# END!
-----------------
"""
\end{lstlisting}

\subsection{Self-Training}
\label{appendix:self-training}

Initially, KL loss is not incorporated, causing the policy model to lose its generalization ability rapidly, despite a continuous decrease in evaluation loss. To address this issue, we introduce KL loss to regularize training from the reference model.  

The logarithmic scaling and weighting factor $\lambda$ are added to balance the impact of KL divergence. Without these adjustments, KL loss would range from 0 to 20000, while the language modeling loss remains between 0 and 12, leading to an imbalance. The logarithm ensures a more stable contribution of KL loss during training.  

As illustrated in Fig.~\ref{fig:KL}, setting $\lambda = 0.5$ achieves a balanced trade-off between KL loss and language modeling loss, preventing excessive divergence from the reference model while ensuring stable and effective training.

% \begin{figure}[t]
% \centering
% \includegraphics[width=1\textwidth]{imgs/kl.png}
%  \caption{Loss dynamics during training. The left plot shows log KL loss and the right plot depicts the causal language modeling loss. By setting $\lambda = 0.5$, the log KL loss and language modeling loss remain relatively balanced throughout training.}

%  \label{fig:KL}
% \end{figure}

\end{document}